# Unraveling the Size Determination Mechanism of Nanocrystal Synthesis via Interpretable Neural Networks

Kai Gu[1,*] , Haizheng Zhong[1,*]
[1]MIIT Key Laboratory for Low-Dimensional Quantum Structure and Devices, School of Materials Sciences & Engineering, Beijing Institute of Technology, Beijing 100081, China
e-mail: kaigu@bit.edu.cn; hzzhong@bit.edu.cn

## Abstract

Deep learning models of nanocrystal synthesis enable the prediction of size and shape by encoding precursors and reaction conditions. However, their black-box nature hinders gaining deep insights into the underlying synthetic mechanisms. Here, we develop the Nanocrystal Equation Learner (NanoEQL), a fully white-box neural network to unravel the size determination mechanisms of nanocrystal synthesis. Building on the EQL architecture, eight operators are introduced to replace standard activation functions to fit the mathematical equations in nanocrystal synthesis. Among these operators, three smoothed operators address the gradient explosion of singular operators at zero. To evaluate the weights of different precursors, we develop a temperature-gated attention pooling strategy that encodes concentration-driven and reactivity-driven chemical synthesis mechanisms into the temperature gate. The NanoEQL model illustrates that the final nanocrystal size can be described by a linear equation composed of three scalars representing nanocrystallization capability ($-Z_p$), growth capability ($Z_{rea}$), and external input potential ($-Z_{ops}$). These interpretable scalars not only advance the rational design of nanocrystal synthesis but also establish a generalizable paradigm for deciphering chemical reaction mechanisms through white-box machine learning.

## Introduction

Colloidal nanocrystals are essential nanomaterials synthesized via solution-phase methods, with widespread applications in fields such as optoelectronic devices[1-3], catalysis[4] and bioimaging[5]. Nanocrystal synthesis involves complex chemical reactions and crystallization processes[6, 7], including the prenucleation[8-10], nucleation[11], growth[12], and postgrowth stages[13]. Synthesis parameters exert profound influences across these stages. For example, the precursor reactivity influences not only prenucleation kinetics but also subsequent nanocrystal growth[14-17]. Deep learning models, with their powerful fitting capabilities, have already established relationships between synthesis parameters and product performance in the synthesis of organic compounds[18], metal-organic frameworks[19], and carbon dots[20, 21]. Similarly, deep learning models have achieved great success in predicting the size and optical spectra of nanocrystals[22-25]. However, due to the black-box nature of deep learning, the underlying prediction process cannot be elucidated, offering limited insight into nanocrystal synthesis mechanisms.

Genetic programming-based algorithms are the most popular white-box architectures that explore mathematical expressions from a given dataset[26], and have demonstrated potential in exploring descriptors for material stability[27] and catalytic activity[28]. For high-dimensional input features, the neural network-based equation learner (EQL) has emerged as a promising white-box architecture[29, 30]. By employing symbolic operations as activation functions, EQL has been validated on simple physical systems and polynomial fitting tasks[31], but has not yet been verified in chemical reaction systems.

In this work, we develop the Nanocrystal Equation Learner (NanoEQL), a fully white-box neural network designed to unravel the size determination mechanisms of nanocrystal synthesis. Building upon the EQL framework, eight operators are introduced to fit the mathematical equations in nanocrystal synthesis, such as identity mapping, squaring, and reciprocals. Inspired by the Michaelis-Menten equation[32], three smoothed operators are designed to address the gradient explosion of singular operators at zero. To compute the weights of different precursors, we develop a temperature-gated attention pooling strategy that encodes concentration-driven and reactivity-driven chemical synthesis mechanisms into the temperature gate. This pooling strategy simulates the chemical reaction process and eliminates the sequence dependency of the precursors. The NanoEQL model achieves a mean absolute percentage error (MAPE) of 0.35 and an $R^2$ of 0.65, significantly outperforming other white-box models and matching the performance of black-box tree-based models. The model demonstrates that nanocrystal size is described by a linear equation composed of three scalars representing nanocrystallization capability ($-Z_p$), growth capability ($Z_{rea}$), and external input potential ($-Z_{ops}$). All of these scalars can be expressed as equations derived from input features to demonstrate interpretability.

## Results and Discussion

### EQL Architecture Based on Gated Attention

**Figure 1a** illustrates the architecture of the NanoEQL model. The dataset comprises synthesis recipes for colloidal nanocrystals, containing the chemical formulas of the nanocrystal products and precursors, as well as the reaction conditions (injection temperature, $T_{inj}$; reaction temperature, $T_{rea}$; and reaction time, t). To ensure feature interpretability, the physicochemical features of the nanocrystal products and inorganic precursors were generated using matminer[33], while the features for the organic precursors were generated via RDKit. As shown in Figure S1, variance inflation factor analysis was employed to select features with weak collinearity, retaining 25% (72 dimensions) of inorganic features and 43% (43 dimensions) of organic features. Each feature set is fed into an independent EQL network (denoted as g). Subsequently, reaction condition features and nanocrystal features directly yield the scalars $Z_{ops}$ and $Z_p$. Precursor features are first dimensionally reduced by EQL networks, yielding 4-dimensional latent representations $Z_{inorg}$ and $Z_{org}$ for inorganic and organic precursors, respectively.

A temperature-gated attention mechanism is employed to compute the weight, $w_j$, of each precursor within a given recipe:

$$w_j = \alpha \cdot \widetilde{m}_j + \beta \cdot \tilde{a}_j \tag{1}$$

where $\tilde{m}_j$ and $\tilde{a}_j$ represent the concentration fraction and the reactivity factor of the $j$-th reactant, respectively. The reactivity factor $\tilde{a}_j$ is derived by mapping precursor and reaction condition features through a multilayer perceptron (MLP), given by $\tilde{a}_j = Softmax\left(MLP\left(\left[Z_{inorg,j} \parallel x_{ops}\right]\right)\right)$. The gating coefficients $\alpha$ and $\beta$ dynamically adjust whether the weights are dominated by concentration or by reactivity, defined as $[\alpha, \beta]^T = Softmax\left(W_{gate}\left[T_{inj}, T_{rea}, t\right]^T + b_{gate}\right)$.

Multi-channel weighted pooling is utilized to simulate the chemical reactions among the reactants. For example, average pooling is computed $F_{mean} = \sum_{j=1}^{N} w_j \cdot Z_{inorg,j}$. The resulting pooled features from inorganic and organic channels are then concatenated and fed into the $g_{rea}$ network to derive the scalar $Z_{rea}$. Finally, the three scalars are combined through a linear top-level predictor to calculate the nanocrystal size.

All independent EQL networks share an identical architecture. **Figure 1b** illustrates the schematic of the product EQL network ($g_p$). Each input feature dimension is mapped through eight operators, including identity, squaring, and reciprocal functions, which suffice to describe fundamental chemical reaction formulae. After passing through three hidden layers, the network outputs the scalar $Z_p$. Because the standard reciprocal, square root and cube root operators exhibit excessively fast gradient updates near zero, the model tends to overuse them, thereby deviating from physical laws. To address this issue, we drew inspiration from the Michaelis-Menten equation (which describes the reaction rate between an enzyme and its substrate) and introduced three smoothed operators. **Figure 1c** shows the equations for these smoothed operators. The reciprocal operator is replaced by $\frac{x}{x^2+c}$. When $x \to 0$, $y \approx \frac{x}{c}$, and when $|x| \gg c$, $y \approx \frac{1}{x}$. The gradient at $x = 0$ is $\frac{1}{c}$. The design and gradient calculations for all operators are summarized in **Table S1**.

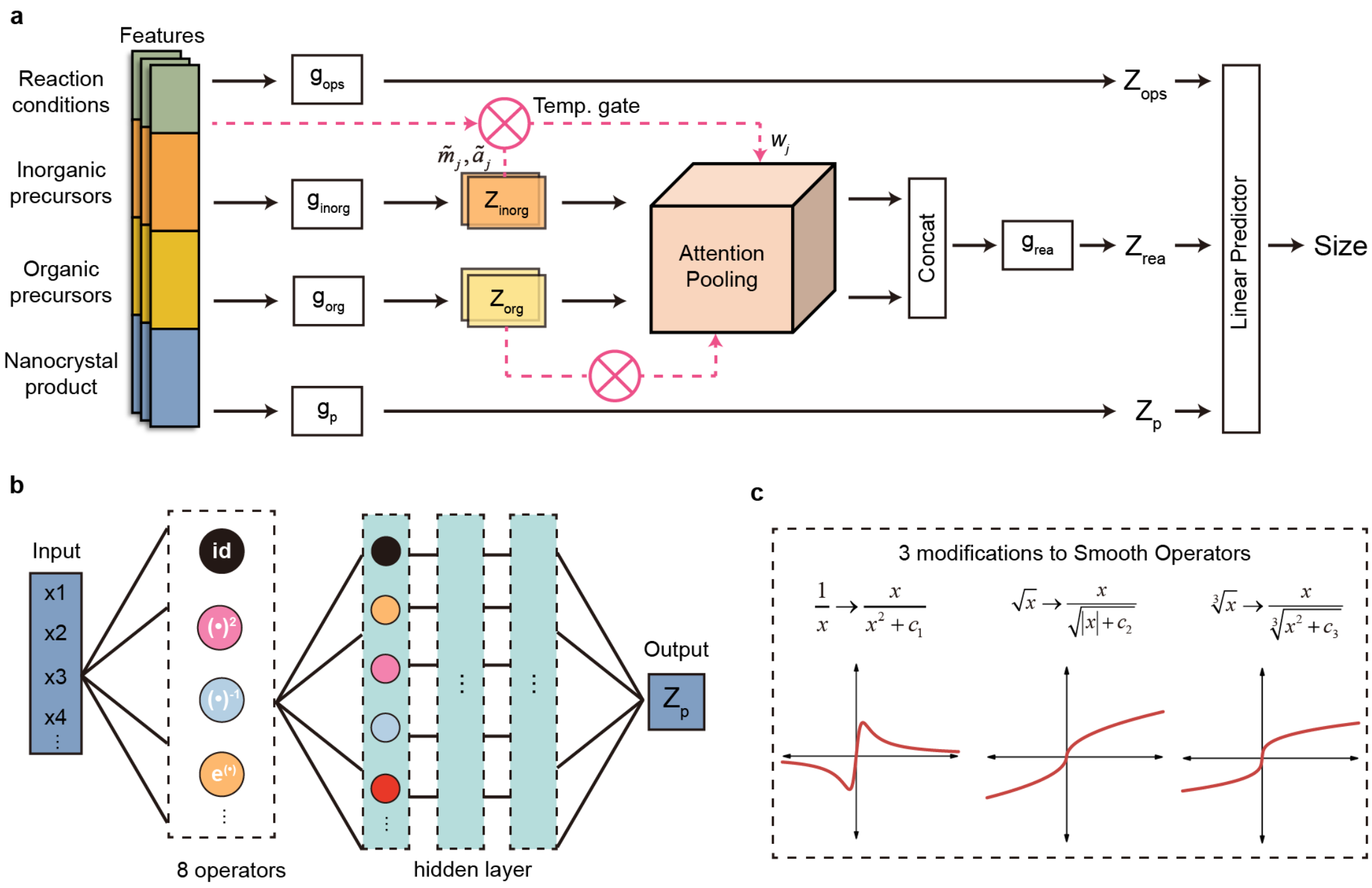

**Figure 1. Model architecture and operator design of NanoEQL. (a)** The overall architecture of NanoEQL, where the features of the reaction conditions and the nanocrystal products directly output scalars $Z_{ops}$ and $Z_p$ via independent EQL networks. Precursor features are first dimensionally reduced, then combined via a gated attention pooling mechanism, concatenated, and processed by the $g_{rea}$ network to yield scalar $Z_{rea}$. Finally, the three scalars are fed into a linear top-level predictor to output the nanocrystal size. **(b)** Schematic illustration of the product EQL network ($g_p$). Here, id denotes identity mapping, $(\cdot)^2$ denotes the square operator, $(\cdot)^{-1}$ denotes the reciprocal operator, and $e^{(\cdot)}$ represents the exponential operator. **(c)** Smoothed modifications for the reciprocal, square root, and cube root operators.

## Performance of NanoEQL

**Figure 2a** shows a performance comparison between NanoEQL and various white-box and black-box models. For nanocrystal size prediction, NanoEQL achieves an $R^2$ of 0.65 and a MAPE of 0.35 on the test set, outperforming other genetic programming-based white-box models such as PySR[34] ($R^2$=0.27). Among black-box models, tree-based models exhibit strong performance. For instance, random forest achieves an $R^2$ of 0.64 and a MAPE of 0.32, a performance level nearly comparable to that of NanoEQL. **Figure 2b** and **2c** show parity plots for size predictions from NanoEQL and random forest, respectively. Both models yield highly accurate predictions below 25 nm. Larger prediction deviations are observed for larger sizes, which can be attributed to the inherent measurement uncertainty associated with larger nanocrystals.

The choice of the top-level predictor in NanoEQL impacts the overall model performance. **Table S2** compares the predictive performance when employing linear, MLP, quadratic, and EQL as the top-level predictor. The linear predictor yields the best results, whereas the more complex MLP and EQL architectures perform the worst ($R^2$=0.6). This indicates that the features are effectively decoupled at the bottom layers of the network, eliminating the need for a complex top-layer structure. Moreover, a simple top-level linear network facilitates more efficient backpropagation and weight updates in the lower layers. In addition, we apply a tenfold higher learning rate to the gating parameters to ensure their activation (**Figure S2**).

Zero-valued features are prevalent in the dataset, which cause abnormally large gradients for the original reciprocal operator near zero. Even with the addition of clamping and shifting strategies (e.g., $\frac{1}{x+0.1}$), the operator outputs a constant term when the feature value is zero, rendering the reciprocal operator ineffective. As shown in **Figure 2d** and **Table S3**, the introduction of the smoothed reciprocal operator successfully prevents the model from abusing this function and improves the predictive performance ($R^2$ increases from 0.54 to 0.65).

The NanoEQL architecture is also applicable to organic chemical reactions. As shown in **Figure 2e**, NanoEQL exhibits competitive performance on the Buchwald-Hartwig cross-coupling reaction dataset[35]. For yield prediction, it achieves an $R^2$ of 0.93 and a mean absolute error (MAE) of 4.89%, exceeding random forest and XGBoost and approaching the performance of LightGBM ($R^2$=0.94 in **Table S4**).

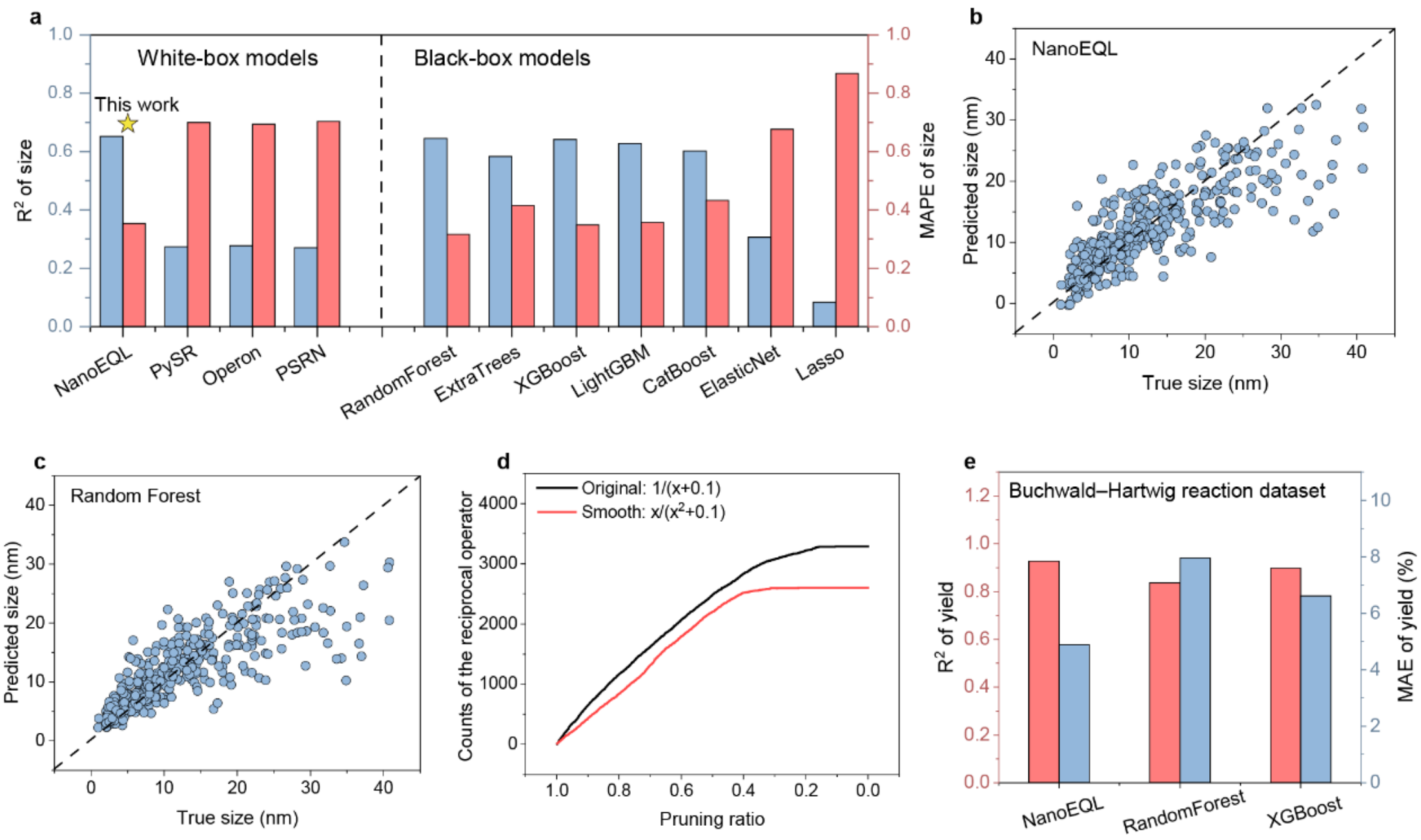

**Figure 2. Performance of NanoEQL. (a)** Performance comparison between NanoEQL and other white-box and black-box models for size prediction; Parity plots of the size predictions on the independent test set for **(b)** NanoEQL and **(c)** the random forest model; **(d)** Usage counts of the reciprocal operator under different pruning ratios; **(e)** Performance comparison of NanoEQL with random forest and XGBoost in predicting the yield of organic reactions.

## Interpretability of NanoEQL

NanoEQL reveals that nanocrystal size follows a simple linear relationship with the three scalars $Z_p$, $Z_{ops}$, and $Z_{rea}$, as given in Equation (2). The relative magnitudes of the coefficients indicate that reaction features are more important than product features, which in turn outweigh reaction condition features.

$$Size = 5.344 \cdot Z_{rea} + 2.245 \cdot (-Z_{ops}) + 2.53 \cdot (-Z_p) + 14.97 \quad (2)$$

Owing to the regularization effect, NanoEQL exhibits excellent sparsity (**Figure S3**), allowing each scalar to be represented by an equation composed of the original input features. **Figure 3** shows the computational graph of the pruned network for each scalar, where neuron colors represent different operators. At a pruning ratio of 60%, the model performance on both training and test sets reaches a plateau (**Figure S4a and 4b**). However, the complexity of the equations increases exponentially as the pruning ratio decreases. For instance, at a pruning ratio of 1%, the complexity of the $Z_{rea}$ equation reaches $10^5$ (**Figure S4c**).

As shown in **Figure 3a**, when the pruning ratio of $g_p$ is 99.5%, $Z_p$ can be represented by the following system of equations (3):

$$\begin{cases} Z_p = 0.1058 \cdot \dfrac{y_3}{{y_3}^2 + 0.1} \\ y_3 = -0.0613 \cdot {y_{2,1}}^2 + 0.0741 \cdot \dfrac{y_{2,2}}{{y_{2,2}}^2 + 0.1} - 0.0695 \cdot \dfrac{y_{2,3}}{{y_{2,3}}^2 + 0.1} \\ y_{2,1} = 0.0392 \cdot x_1 - 0.0522 \cdot (e^{x_2} - 1) \\ y_{2,2} = 0.0418 \cdot \dfrac{x_3}{{x_3}^2 + 0.1} + 0.0414 \cdot (e^{x_4} - 1) \\ y_{2,3} = 0.04 \cdot x_1 - 0.0389 \cdot {x_5}^2 \end{cases} \tag{3}$$

where $x_1$ denotes the feature mean NfValence, defined as the composition-weighted mean value of the number of f-orbital valence electrons computed across the constituent elements, and $x_2$ denotes maximum m_p, defined as the maximum value among the constituent elements of the p-type effective mass computed across the constituent elements. Descriptions of the remaining features are summarized in **Table S5**. The $g_p$ network focuses primarily on the positions of the constituent elements in the periodic table. For example, $x_1$ serves to identify the presence of rare-earth elements, while $x_2$ identifies the presence of heavy-hole elements such as O, F, and N. **Figure 3b** shows the computational graph of $g_{ops}$ at a pruning ratio of 99.2%. Because there are only three input features, the regularization effect becomes active at the second hidden layer (**Figure S3a**).

**Figure 3c** shows the computational graphs of the pruned $g_{inorg}$, $g_{org}$, and $g_{rea}$ networks. In contrast to the product network, the inorganic precursor network focuses on intrinsic physicochemical properties, such as thermal conductivity, electronegativity, d-orbital electron concentration, and shear modulus. The organic precursor network emphasizes surface electron distribution and polarizability.

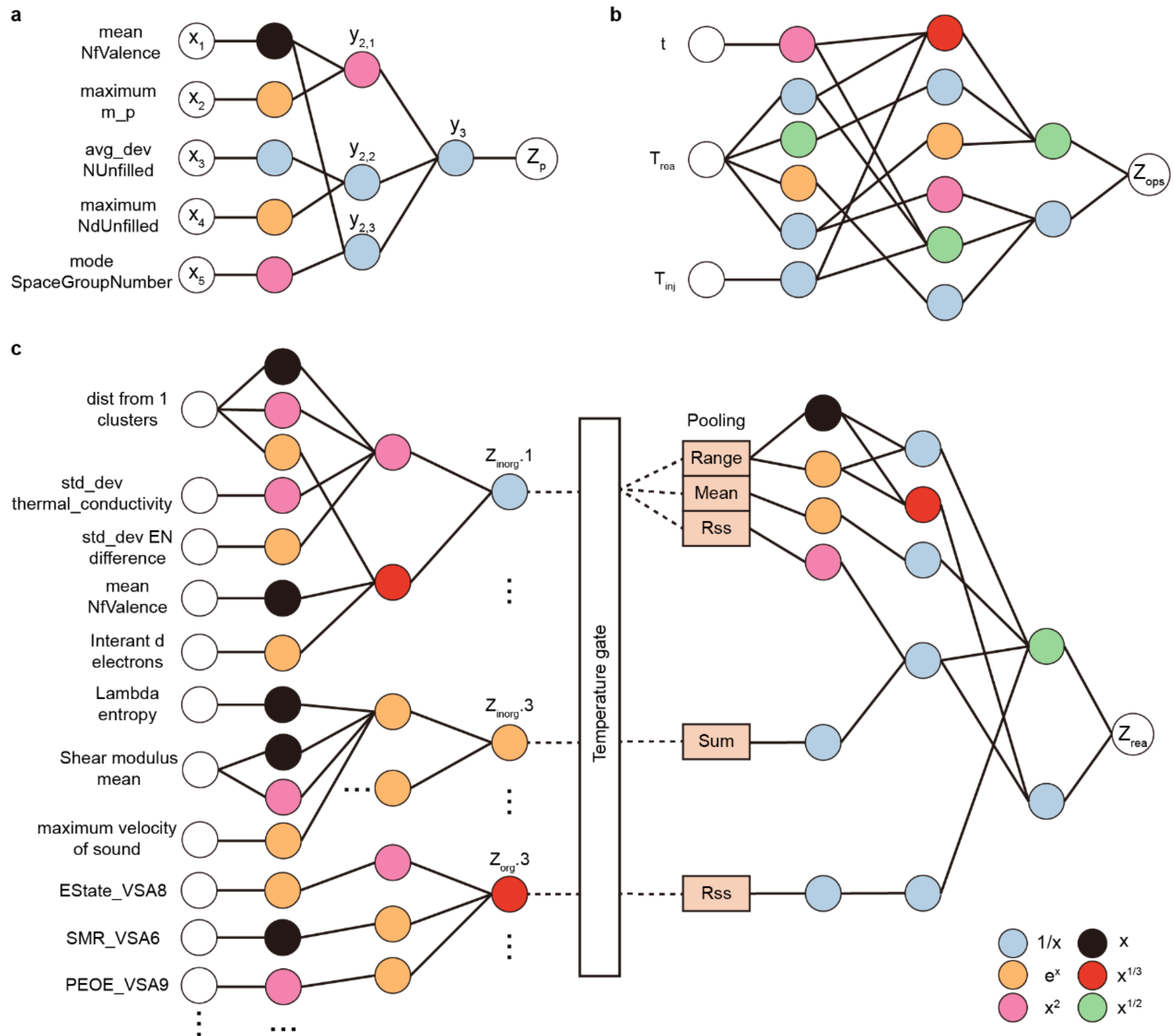

**Figure 3. Computational graphs of the pruned EQL networks, where neuron colors represent different operators. (a)** Computational graph of the $g_p$ network at a pruning ratio of 99.5%; **(b)** Computational graph of the $g_{ops}$ network at a pruning ratio of 99.2%; **(c)** Computational graphs of the $g_{inorg}$, $g_{org}$, and $g_{rea}$ networks, with pruning ratios of 99.3%, 99.6%, and 99.6%, respectively. $Z_{inorg}.1$ represents the first dimension in the four-dimensional inorganic precursor features.

### Physical Significance of the Three Scalars and Synthesis Mechanisms

Based on the NanoEQL architecture, $Z_p$ reflects the intrinsic properties of the product nanocrystals. According to Equation (2), we define $-Z_p$ as the nanocrystallization capability, where a smaller value indicates a higher propensity for nanocrystallization. Luminescent quantum dots represent an important class of nanocrystal materials. We screened 6,800 direct-bandgap semiconductors (0.5~3 eV) and computed their $-Z_p$ values to evaluate their nanocrystallization potential. **Figure 4a** shows the ranking of $-Z_p$ for these materials. $InSnCl_3$, $MoTe_2$, InP, $MgIn_2O_4$, and ZnSe exhibit low $-Z_p$ values of -3.97, -3.94, -2.79, -2.71, and -2.66, respectively. Among these, $MoTe_2$, InP, and ZnSe have all been reported as nanocrystals, with the latter two being commercially promising quantum dots for display[36, 37]. $MnH_4(OF_2)_2$, $Cs_2NaVSe_4$, $K_3NaFeO_4$, and $Cs_3MnBr_5$ show large $-Z_p$ values of 4.27, 3.99, 3.41, and 3.11, respectively, indicating their low potential for nanocrystallization. Given that metal nanocrystals represent another highly active research field, we also calculated the $-Z_p$ values for 34,000 metallic materials (**Figure S5**). **Figure 4b** illustrates the potential energy surface distribution of $-Z_{ops}$. Higher reaction temperatures and longer reaction times lead to larger nanocrystals, while the injection temperature also slightly affects the potential surface distribution (**Figure S6**).

In nanocrystal synthesis, precursor concentration is closely related to the number of nuclei and their subsequent growth, while precursor reactivity governs the reaction rate[14]. Reaction conditions, such as temperature, exert a broader influence. We extracted the temperature-gated parameters from the model to elucidate the underlying synthesis mechanisms. **Figure 4c** shows the distribution of the reactivity coefficients ($\beta$) for both inorganic and organic precursors. The reactivity coefficients are initialized to 0.28 before training. For inorganic precursors, the average reactivity coefficient is 0.18 after training, indicating that their reactions are predominantly concentration-driven. They exhibit reactivity-driven behavior only at high temperatures (>300 °C) (**Figure 4d**). For organic precursors, the mean reactivity coefficient is 0.33, indicating a higher dependence on reactivity compared to inorganic precursors. Concentration-driven behavior is observed only at very short reaction times (**Figure 4e**). We further compared the temperature-dependent reactivity variations of common ligands and solvents. Oleylamine exhibits substantial changes in reactivity (**Figure 4f**). In contrast, oleic acid demonstrates high reactivity exclusively at high temperatures, trioctylphosphine shows high reactivity during prolonged reactions, and the solvent octadecene shows minor reactivity changes (**Figure S7**).

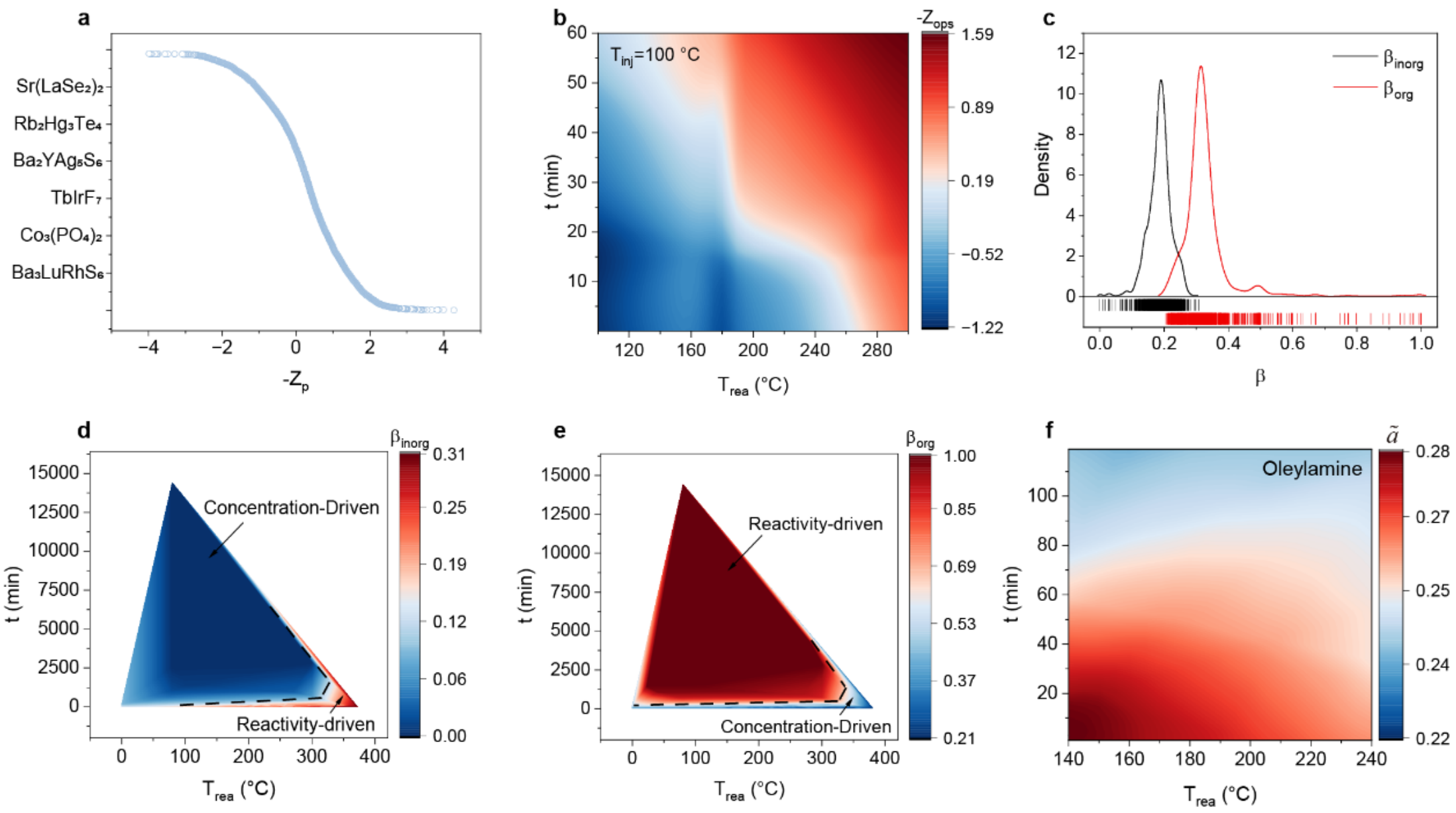

**Figure 4. Mechanistic interpretation based on the temperature gate. (a)** Ranking of -$Z_p$ for 6,800 screened direct-bandgap semiconductors; **(b)** Evolution of -$Z_{ops}$ with reaction temperature and reaction time; **(c)** Distribution of the reactivity coefficients; Evolution of the reactivity coefficients for **(d)** inorganic and **(e)** organic precursors as a function of reaction temperature and time; **(f)** Evolution of oleylamine reactivity with respect to reaction temperature and time.

We reduced the concatenated 56-dimensional pooled precursor features to two dimensions to map the chemical reaction space. Taking the synthesis of PbSe nanocrystals as a case study, **Figure 5a** shows the chemical reaction space and the $Z_{rea}$ distribution for 180,000 samples. Samples with high $Z_{rea}$ values are distributed on the left side of the reaction space, whereas those with low $Z_{rea}$ values are clustered on the right. According to Equation (2), $Z_{rea}$ is proportional to nanocrystal size. Therefore, $Z_{rea}$ represents the growth capability of the nanocrystals. We independently varied $T_{rea}$, t, the Pb precursor content, and the Se precursor content, generating four trajectories within the chemical reaction space (indicated by green dashed boxes). Magnified views of trajectories 1 and 2 are shown in **Figure 5b** and **5c**, demonstrating that increases in $T_{rea}$ and t have a minimal effect on $Z_{rea}$. Trajectories 3 and 4 span nearly half of the reaction space (**Figure 5d and 5e**). As the Pb content increases, $Z_{rea}$ increases, resulting in a larger nanocrystal size. Conversely, an increase in the Se content leads to a decrease in $Z_{rea}$, which reduces the nanocrystal size. To verify this finding, a series of PbSe nanocrystals was synthesized using different Se contents, while all other reaction conditions were kept constant. **Figures 5f** to **5i** show their transmission electron microscopy (TEM) images. The average size of the resulting PbSe nanocrystals is 9.5 nm when the Pb:Se ratio is 1:0.25, while it is 6.1 nm when the Pb:Se ratio is 1:4.

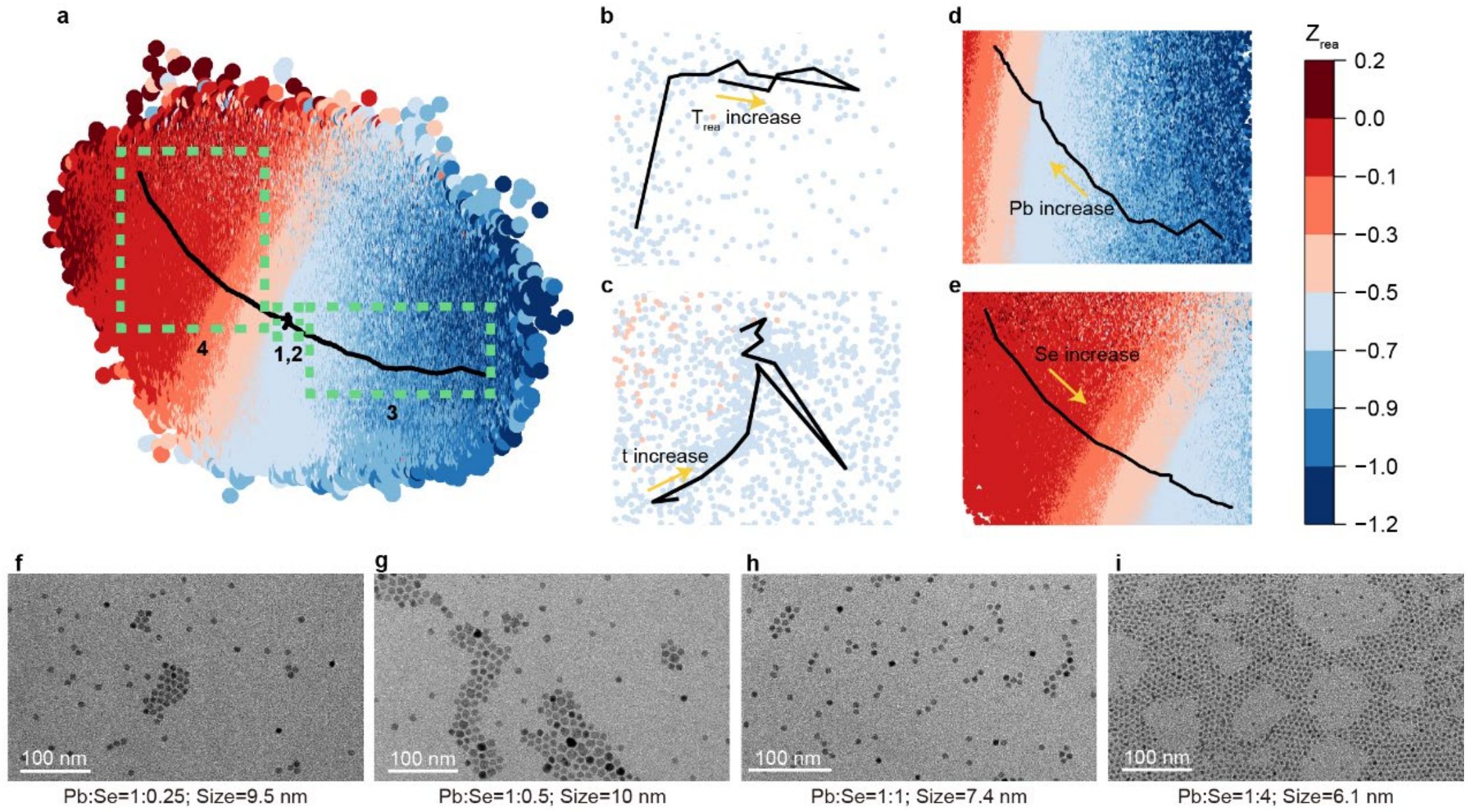

**Figure 5. Chemical reaction space for PbSe nanocrystals. (a)** Chemical reaction space and $Z_{rea}$ distribution of 180,000 PbSe nanocrystal samples. The numbered green dashed boxes (in ascending order) represent trajectories generated by independently varying $T_{rea}$, t, Pb content, and Se content, respectively. Magnified views of the trajectories corresponding to changes in **(b)** $T_{rea}$, **(c)** t, **(d)** Pb content, and **(e)** Se content. TEM images of PbSe nanocrystals with Pb:Se ratios of **(f)** 1:0.25, **(g)** 1:0.5, **(h)** 1:1, and **(i)** 1:4.

## Conclusions

In summary, we develop the Nanocrystal Equation Learner (NanoEQL), a fully white-box neural network designed to unravel the size determination mechanisms of nanocrystals. In the network design, we design three smoothed operators to address the gradient explosion of singular operators at zero. The smoothed operators not only ensure physical interpretability but also improve predictive accuracy by preventing the network from exploiting pathological gradients. To facilitate a deep understanding of nanocrystal synthesis mechanisms, we propose a temperature-gated attention pooling mechanism. This approach encodes concentration-driven and reactivity-driven synthesis mechanisms into the temperature gate to evaluate the weights of various precursors. The pooling operation simulates the chemical reaction process while eliminating the sequence dependency of the precursors. The predictive performance of NanoEQL significantly outperforms that of other white-box models and matches that of black-box tree-based models, achieving a mean absolute percentage error (MAPE) of 0.35 and an $R^2$ of 0.65. Beyond nanocrystal synthesis, the NanoEQL architecture also demonstrates strong predictive power in organic reaction systems, achieving an $R^2$ of 0.93 on the Buchwald-Hartwig reaction dataset for yield prediction, which highlights the generality of the proposed framework.

We focus on interpreting the trained model to elucidate nanocrystal synthesis mechanisms. The results demonstrate that the nanocrystal size can be described by a simple linear equation composed of three scalars: $-Z_p$, $Z_{rea}$, and $-Z_{ops}$. These scalars hold significant physical meaning, representing the nanocrystallization capability, the growth capability, and the external input potential, respectively. Crucially, they can be expressed through a system of equations derived from the original input features. Because $-Z_p$ is independent of the reactants and reaction conditions, it can evaluate the nanocrystallization capability of a given material. We have demonstrated its practical utility through examples screening potential quantum dots and metal nanocrystals. Finally, an analysis of the temperature gating parameters reveals that the average reactivity coefficients for inorganic and organic precursors are 0.18 and 0.33, respectively. This indicates that the reactions of inorganic precursors are predominantly concentration-driven, whereas the reactions of organic precursors exhibit a higher dependence on reactivity.

## Methods

### Dataset and Feature Construction.

The nanocrystal synthesis dataset was obtained from our previous work[25],

comprising solution-phase synthesis recipes together with the size and shape of the resulting products. The Buchwald-Hartwig reaction dataset was taken from the literature[35]. The target columns for the two datasets are size and yield, respectively. Data without valid descriptors was removed. 15% of the data was reserved as the test set using stratified sampling based on the target column. Chemical formulas and three-dimensional structures of the product nanocrystals and inorganic precursors were featurized using matminer[33]. Organic precursors were featurized using RDKit. All descriptors were screened using the Variance Inflation Factor (VIF), retaining only those with $VIF<10$. The shape of the nanocrystals was mapped to three-dimensional features (circularity, aspect ratio, vertices) and concatenated with the product features.

**Model Architecture and Training**

The model input comprises three categories of features: (i) product-level features; (ii) inorganic and organic descriptors of the precursors together with their molar amounts; and (iii) three reaction operation descriptors, namely the injection temperature ($T_{inj}$), reaction temperature ($T_{rea}$), and reaction time (t). For organic reactions, this framework still applies, except that the reaction operation descriptors are replaced by reaction temperature, reaction time, whether the reaction takes place in an inert atmosphere, and whether it is a closed system. At the operator level, each EQL layer invokes a predefined function library to expand the input features. The library includes identity mapping, squaring, smoothed square root, smoothed cube root, smoothed reciprocal, exponential, natural logarithm, and common logarithm. The transformed features are subsequently recombined through linear operations.

For organic and inorganic precursor features, we adopted a design paradigm of dimensional reduction followed by pooling. Independent early EQL sub-networks first compress the high-dimensional features into low-dimensional embedding vectors. These embeddings are then aggregated by a multi-channel pooling module equipped with an attention mechanism. For each precursor group, the pooling module combines the descriptor embeddings with the process descriptors to estimate reactivity-based importance weights through a small attention scorer. Simultaneously, the molar amounts of the precursors are normalized to form mass-based weights. A temperature-dependent gating layer, driven by the injection temperature, reaction temperature, and reaction time, generates two coefficients via a softmax function to balance the mass and reactivity contributions. Based on these hybrid weights, the model computes 7 pooling statistics for both the inorganic and organic precursor sets, including weighted sum, weighted mean, range, weighted standard deviation, square root of the weighted sum of squares, maximum, and minimum. The inorganic and organic pooling statistics are concatenated and fed into an EQL network to generate the variable $Z_{rea}$. The final prediction is obtained by concatenating the outputs from all active branches and passing them through the top-level predictor.

The optimization objective combines a Smooth $L_1$ regression loss with an $L_1$ penalty applied exclusively to the weights and biases of the EQL layers:

$$Loss = SmoothL_1(\hat{y}, y) + \lambda_{L1} \sum \left\| \theta_{EQL} \right\|_1 \tag{4}$$

This design encourages sparsity in the symbolic layers while preventing the attention scorer from collapsing under direct regularization. Parameters are optimized using AdamW. The gating coefficients α and β are initialized to approximately 0.72 and 0.28, respectively, biasing the model toward concentration-driven precursor weighting at the outset of training. To prevent the gate from remaining dormant, the attention-gate parameters are grouped separately and assigned a higher learning-rate multiplier relative to the remaining network parameters. This multiplier is treated as a hyperparameter and optimized via Optuna, allowing the mass-reactivity balance to adapt more rapidly. A cosine-annealing warm-restart scheduler is applied during training, and gradient norms are clipped at each iteration to stabilize optimization.

Hyperparameter selection is performed using the Optuna framework. Each Optuna trial is evaluated by five-fold cross-validation on the training set. After the hyperparameters are determined, the final training epoch is selected via out-of-fold validation, choosing the epoch with the lowest average mean squared error across the five folds (**Figure S8**). With the training epoch and hyperparameters fixed, the model is evaluated on the independent test set. The test set is never accessed during hyperparameter tuning or epoch selection. All experiments were conducted on a server equipped with an NVIDIA RTX 4090 GPU.

**Screening of Materials for -$Z_p$ Calculation**

Semiconductor materials were retrieved from the Materials Project[38] and screened according to the following criteria: (i) exclusion of actinide elements; (ii) bandgap between 0.3 and 3 eV; (iii) direct bandgap; (iv) nonmetallic; (v) fewer than four constituent elements; (vi) fewer than 50 sites in the unit cell; and (vii) energy above the convex hull less than 0.05 eV atom$^{-1}$. This screening yielded 6,800 semiconductor materials. Metallic materials were screened using criteria (i), (v), (vi), and (vii), resulting in 34,000 metallic materials.

**Construction of the PbSe Nanocrystal Reaction Space**

The reaction recipes for PbSe nanocrystals were uniformly generated by establishing boundaries for each independent input variable. $T_{inj}$ was set strictly equal to $T_{rea}$, with a range of 140 to 300 °C, and t ranged from 1 to 120 min. The molar amounts of PbO and Se ranged from 1 to 10 mmol; the molar amounts of oleylamine, oleic acid, and trioctylphosphine ranged from 4 to 40 mmol; and the molar amount of 1-octadecene ranged from 4 to 100 mmol. The 56-dimensional reaction space was reduced to two dimensions using the UMAP algorithm. The 56-dimensional feature vector is formed by concatenating the seven pooling statistics computed over the 4-dimensional inorganic-precursor embeddings ($Z_{inorg}$) with the seven pooling statistics computed over the 4-dimensional organic-precursor embeddings ($Z_{org}$).

**Synthesis and Characterization of PbSe Nanocrystals**

The hot-injection synthesis method and chemical reagents are detailed in the reference[39]. Lead oxide, octadecene, oleylamine, and oleic acid were fixed at 5 mmol, 46.9 mmol, 5 mmol, and 22.2 mmol, respectively, while the amount of selenium powder varied from 1.25 mmol to 20 mmol (the molar ratio of tri-n-octylphosphine to selenium powder was maintained at 1.21 throughout). The reaction temperature and reaction time were set at 160 °C and 10 min, respectively. All nanocrystals were not subjected to any size selection prior to TEM characterization. TEM observations were performed using a FEI Tools F200S (FEI Co., USA) field-emission transmission electron microscope operated at 200 kV and 120 kV, respectively. The average size of nanocrystals was obtained by undifferentiated counting of hundreds of particles.

**Data availability**

All data are provided in the main text or Supporting Information. The dataset and training code for NanoEQL are available at https://github.com/ime1452/Nanocrystal-Equation-Learner.

**Supporting Information**

Additional figures and tables that provide detailed insights into various aspects of the study, including comparison of formulas and gradients for the eight operators (Table S1); Comparison of different top-level predictors (Table S2); comparison of different reciprocal operators (Table S3); performance comparison of different models in the yield prediction task (Table S4); feature explanations (Table S5); a schematic diagram of feature engineering for the dataset (Figure S1); curve of the gate factor varying with training epochs (Figure S2); weight distributions in the hidden layers of the $g_p$ and $g_{ops}$ networks (Figure S3); evolution of network's performance on the training and test sets and its equation complexity under different pruning ratios (Figure S4); $-Z_p$ sorting of 34,000 metal materials and some typical examples (Figure S5); potential energy surface distributions for $-Z_{ops}$ (Figure S6); evolution of the reactivity of oleic acid, trioctylphosphine, and octadecene with reaction temperature and reaction time (Figure S7); determine the training epoch based on the lowest average MSE obtained through 5-fold cross-validation (Figure S8).

## Acknowledgements

This work is granted by the National Natural Science Foundation of China (Nos. 52525309, and U23A20683) and the Beijing Municipal Science & Technology Commission, Administrative Commission of Zhongguancun Science under Park No. Z231100006023018.

## Author information

Authors and Affiliations

MIIT Key Laboratory for Low-Dimensional Quantum Structure and Devices, School of Materials Sciences & Engineering, Beijing Institute of Technology, Beijing 100081, China

Kai Gu, Haizheng Zhong

Contributions

K. G., and H. Z. conceived the project. K. G. synthesized and characterized the nanocrystals. K. G. performed the model training and evaluation. K. G. and H. Z. analyzed the models and wrote the manuscript.

Corresponding authors

Correspondence to Kai Gu or Haizheng Zhong

**Notes**

The authors declare no competing interests.

**Table of Contents**

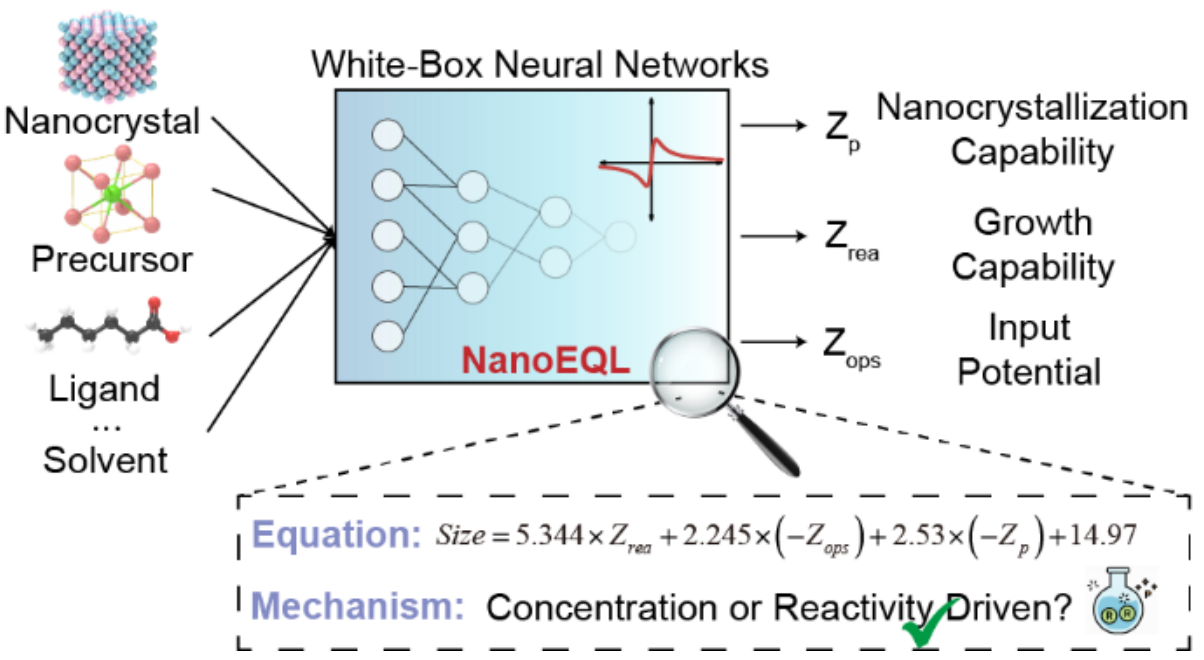